\documentclass{article}
\usepackage[utf8]{inputenc}
\usepackage[table,xcdraw]{xcolor}
\usepackage{authblk}
\usepackage{graphicx}
\usepackage{amsmath}
\usepackage[a4paper, total={6in, 8in}]{geometry}
\usepackage{xspace}
\newcommand{\model}{VISIOCITY\xspace}
\usepackage{caption,subcaption}
\usepackage{multirow}

\title{How Good is a Video Summary? A New Benchmarking Dataset and Evaluation Framework Towards Realistic Video Summarization}

\author[1]{Vishal Kaushal}
\author[2]{Suraj Kothawade}
\author[1]{Anshul Tomar}
\author[2]{Rishabh Iyer}
\author[1]{Ganesh Ramakrishnan}
\affil[1]{Department of Computer Science, Indian Institute of Technology Bombay}
\affil[2]{Department of Computer Science, University of Texas at Dallas}

\date{}
      
\begin{document}

\maketitle

\begin{abstract}
Automatic video summarization has attracted a lot of interest. However it is still an unsolved problem due to several challenges. We take steps towards making automatic video summarization more realistic by addressing the following challenges. The currently available datasets either have very short videos or have few long videos of only a particular type. We introduce a new benchmarking video dataset called \model (VIdeo SummarIzatiOn based on Continuity, Intent and DiversiTY) which comprises of longer videos across six different categories with dense concept annotations capable of supporting different flavors of video summarization and other vision problems. Secondly, for long videos, human reference summaries necessary for supervised video summarization techniques are difficult to obtain. We explore strategies to automatically generate multiple reference summaries from indirect ground truth present in \model. We show that these summaries are at par with human summaries. We also present a study of different desired characteristics of a good summary and demonstrate how, especially in long videos, it is quite possible and frequent to have two good summaries with different characteristics. Thus we argue that evaluating a summary against one or more human summaries and using a single measure has its shortcomings. We propose an evaluation framework for better quantitative assessment of summary quality which is closer to human judgment. Lastly, we present insights into how a model can be enhanced to yield better summaries. Sepcifically, when multiple diverse ground truth summaries can exist, learning from them individually and using a combination of loss functions measuring different characteristics is better than learning from a single combined (oracle) ground truth summary using a single loss function. We demonstrate the effectiveness of doing so as compared to some of the representative state of the art techniques tested on \model. We release \model as a benchmarking dataset and invite researchers to test the effectiveness of their video summarization algorithms on \model.
\end{abstract}

\section{Introduction and Motivation}
The unprecedented rise in the amount of video data has also made it difficult to consume them. This has given rise to the need for automatic video summarization techniques which aim at producing much shorter videos without significantly compromising on the key information contained in them. Consequently there has been a lot of work pushing the state-of-the-art for newer algorithms and model architectures~\cite{fu2019attentive, ji2019video, fajtl2018summarizing, yuan2019cycle, zhang2016video, zhang2016summary, zhou2018deep, gygli2015video, xiong2019less, li2018local} and datasets~\cite{gygli2014creating, potapov2014category, song2015tvsum}. However the literature also talks of a few fundamental challenges that need to be addressed before we have a more realistic video summarization that works in practice. \textbf{In this work we take steps towards addressing the following challenges:}

\textbf{Dataset}: For a true comparison between different techniques, a benchmark dataset is critical. Almost all recent techniques have reported their results on TVSum~\cite{song2015tvsum} and SumMe~\cite{gygli2014creating} which have emerged as benchmarking datasets of sorts. However, since the average video length in these datasets is of the order of \emph{only} 1-5 minutes, they are far from being effective in real-world settings characterized by long videos. While there have been several attempts at creating datasets for video summarization, \textbf{they either a) have very short videos, or b) they have very few long videos and often of only a particular type.} A large dataset with a lot of different types of full-length videos with rich annotations to be able to support different techniques was one of the recommendations in~\cite{truong2007video}, is still not a reality and is clearly a need of the hour~\cite{ji2019video}. \textbf{We introduce \model to address this need.} \model is a diverse collection of 67 long videos spanning across six different categories with dense concept annotations. Furthermore, different flavors of video summarization, for example, query focused video summarization~\cite{xiao2020convolutional, vasudevan2017query, sharghi2017query}, are often treated differently and on different datasets. With its rich annotations \model can lend itself well to other flavors of video summarization and also other computer vision video analysis tasks like captioning or action recognition. Also, since the videos span across different well-defined domains, \model is suitable for more in-depth domain specific studies on video summarization~\cite{truong2007video, potapov2014category}.

\textbf{Reference summaries for supervised learning}: Supervised techniques tend to work better than unsupervised techniques~\cite{ji2019video, zhang2016video} because of learning directly from human summaries. Video summaries are highly context dependent (depends on the purpose behind getting a video summary), subjective (even for the same purpose, preferences of two persons don't match) and depends on high-level semantics of the video (two visually different scenes could capture the same semantics or visually similar looking scenes could capture different semantics). As an example of context, one may want to summarize a surveillance video either to see a 'gist' of what all happened or to quickly spot any 'abnormal' activity. As an example of personal preferences or subjectivity, while summarizing a {\it Friends} video (a popular TV series), two users may have different opinion on what is 'important' or 'interesting'. Similar example for higher level semantics is that closeup of a player in soccer can be considered important if it is immediately followed by a goal, while not so important when it occurs elsewhere, even though both look visually the same. \textbf{Thus there is no single 'right' answer and two human summaries could be quite different in their selections}~\cite{kannappan2019human, otani2019rethinking}. In a race to achieve better performance, most state of the art techniques are based on deep architectures and are thus data hungry. The larger the dataset and more the number of reference summaries to learn from, the better. \textbf{Unfortunately, for long videos getting human summaries is very time consuming. It becomes increasingly expensive and, beyond a point, infeasible to get these reference summaries from humans. Also, this is not scalable to experiments where reference summaries of different lengths are desired}~\cite{gygli2015video}. To alleviate this problem, in this work \textbf{we explore strategies to automatically generate ground truth reference summaries which can be used to train a model}.

Further, most supervised learning approaches are trained using a 'combined' ground truth summary, either in form of combined scores from multiple ground truth summaries or scores~\cite{zhang2016summary,fajtl2018summarizing, ji2019video} or in form a set of ground truth selections, as in dPPLSTM~\cite{zhang2016summary}. However, combining them into one misses out on the separate flavors captured by each of them. Combining many into one set of scores also runs the risk of giving more emphasis to 'importance' over and above other desirable characteristics of a summary like continuity, diversity etc. This is also noted by~\cite{apostolidis2020unsupervised, zhou2018deep} where they argue that \textbf{supervised learning approaches, which rely on the use of a combined ground-truth summary, cannot fully explore the learning potential of such architectures.} The necessity to deal with different kind of summaries in different ways was also observed by~\cite{truong2007video}.~\cite{apostolidis2020unsupervised, zhou2018deep} use this argument to advocate the use of unsupervised approaches. \textbf{We leverage \model to demonstrate that better results can be achieved when a supervised model learns from individual ground truth summaries using multiple loss functions each measuring deviation from different desired characteristics of summaries.}


\textbf{Evaluation}: A video summary is typically evaluated by comparing it against human summaries, for example using F1 score defined as harmonic mean of precision (ratio of temporal overlap between candidate and reference summary to duration of summary) and recall (ratio of temporal overlap between candidate and reference summary to video duration)~\cite{zhou2014learning, fajtl2018summarizing, ji2019video}. To accommodate multiple human summaries, either average F1 or max F1~\cite{song2015tvsum, gygli2015video} is reported. However, \textbf{a good candidate may get a low score just because it was not fortunate to have a matching human summary, a likely scenario in case of long videos.} Furthermore, F1 score has some limitations. Due to the segmentation used as a post processing step in typical video summarization pipeline, even random summaries can get good F1 scores~\cite{otani2019rethinking}. Also F1 is not designed to measure aspects like continuity and diversity of a summary. Two summaries may have same F1 score, and yet one may be more continuous (and hence visually more pleasurable to watch) than another. \textbf{We propose an evaluation framework where a summary is assessed on its own merit using the rich annotations in \model(as against comparing it against available human summaries) using a suite of measures to capture various aspects of a summary like continuity, diversity, redundancy, importance etc. (as against over dependence on one measure).}

\section{Related Work}

\subsection{Datasets} Currently available datasets for video summarization either have very short videos or have few long videos of only a particular type. Table~\ref{tab:datasets} compares \model with other existing datasets for video summarization. MED Summaries Dataset~\cite{potapov2014category} consists of 160 annotated videos of length 1-5 minutes, with 15 event categories like birthday, wedding, feeding etc. The annotation comprises of segments and their importance scores. TVSum~\cite{conf/cvpr/SongVSJ15} consists of 50 videos (average length 4 minutes) from 10 categories with importance scores provided by 20 annotators for each 2 second snippet. The videos correspond to very short events like `changing vehicle tires`, `making sandwich` etc. Though number of categories in TVSum and MedSummaries appear to be large, the notion of categories there is of events, like ‘making a sandwich’ or ‘attempting bike tricks’, quite different from different \emph{domains} in VISIOCITY with an intent of studying the characteristics of summaries of different types of videos like sports or TV Shows. The UT Egocentric Dataset~\cite{lee2012discovering}  consists of long and annotated videos captured from head mounted cameras. However, though each video is very long, there are only 4 videos and they are of one type, i.e. egocentric. SumMe~\cite{GygliECCV14} consists of 25 videos with an average length of about 2 min. The annotation is in form of user summaries of length between 5\% to 15\%. Each video has 15 summaries. The VSUMM dataset~\cite{de2011vsumm} consists of two datasets. Youtube consists of 50 videos (1-10 min) and OVP consists of 50 videos of about 1-4 min from the Open Video Project. Each video has 5 user summaries in the form of set of key frames. Tour20~\cite{panda2017diversity} consists of 140 videos with a total duration of 7 hours and is designed primarily for multi video summarization. It is a collection of videos of a tourist place. The average duration of each video is about 3 mins. TV Episodes dataset~\cite{yeung2014videoset} consists of 4 TV show videos, each of 45 mins. The total duration is 3 hours. A recent dataset, UGSum52~\cite{lei2019framerank} offers 52 videos with 25 user generated summaries each. LOL~\cite{fu2017video} consists of online eSports videos from the League of Legends. It consists of 218 videos with each video being between 30-50 mins. The associated summary videos are 5 - 7 mins long. While this dataset is significantly larger compared to the other datasets, it is limited only to a single domain, i.e. eSports. ~\cite{sharghi2018improving} have extended the UTE dataset to 12 videos and have provided concept annotations, but they are limited to only egocentric videos and do not support any concept hierarchy. The scores annotations, as in TVSum etc. are richer annotations, but are limited only to importance scores. VISIOCITY on the other hand comes with dense concept annotations for each snippet. To the best of our knowledge, VISIOCITY is one of its kind large dataset with many long videos spanning across multiple categories and annotated with rich concept annotations for each snippet.

\begin{table*}[ht]
\small{
\centering
\begin{tabular}{ | c | c | c | c | c | c |}
\hline
Name & \# Videos & Duration of Videos & Total Duration & \# Cat\\
 \hline
  SumMe~\cite{GygliECCV14} & 25 & Avg: 2 min 39 secs & 1.10 hours & - \\
 TVSum~\cite{conf/cvpr/SongVSJ15} & 50 & Avg: 4 min 11 sec & 3.5 hours & 10\\
 MED Summaries~\cite{potapov2014category} & 160 & Dur: 1-5 min, Avg: 2.5 min & 9 hours & 15 \\
 UT Egocentric~\cite{lee2012discovering} & 4 & Avg: 254 min 26 sec & 16.96 hours & 1 \\
 Youtube 1~\cite{de2011vsumm} & 50 & Dur: 1-10 min, Avg: 1 min 39 sec & 1.38 hours & -\\
 Youtube 2~\cite{de2011vsumm} & 50 & Dur: 1-4 min, Avg: 2 min 54 sec & 2.41 hours & -\\
 Tour20~\cite{panda2017diversity} & 140 & Avg: 3 min & 7 hours & - \\
 TV Episodes~\cite{yeung2014videoset} & 4 & Avg: 45 min & 3 hours & 1 \\
 LOL~\cite{fu2017video} & 218 & Dur: 30-50 min & - & 1\\ 
 \textbf{VISIOCITY \ (OURS)} & \textbf{67} & \textbf{Dur: 14-121 mins, Avg: 55 mins} & \textbf{71 hours} & \textbf{5}\\
 \hline
 \end{tabular}
 \vspace{-8pt}
\caption{Comparison of VISIOCITY with other datasets in literature. "-" means the corresponding information was not available. "Dur" stands for Duration and \# Cat is \# of event categories available in a dataset}
\label{tab:datasets}}
\end{table*}

\subsection{Techniques for Automatic Video Summarization} A number of techniques have been proposed to further the state of the art in automatic video summarization. Most video summarization algorithms try to optimize several criteria such as diversity, coverage, importance and representation. Some techniques do this through submodular functions ~\cite{zhang2016video,gygli2015video,Kaushal2019Demystifying,gygli2015video,kaushal2019framework}, some use LSTMs~\cite{zhang2016video}, some use determinantal point processes (DPPs~\cite{kulesza2012determinantal})~\cite{zhang2016video, li2018local, gong2014diverse, sharghi2016query}, some use reinforcement learning~\cite{zhou2018deep, lan2018ffnet} and some use attention models~\cite{ji2019video,fajtl2018summarizing}. ~\cite{fu2019attentive} attempts to address video summarization via attention-aware and adversarial training.

\subsection{Evaluation} Evaluation of video summaries is challenging task owing to the multiple definitions of success. Early approaches~\cite{lu2013story, ma2002user} involved user studies but with the obvious demerit of cost and reproducibility. With a move to automatic evaluation, every new technique of video summarization came with its own evaluation criteria making it difficult to compare results different techniques. Some of the early approaches included VIPER~\cite{doermann2000tools}, which addresses the problem by defining a specific ground truth format which makes it easy to evaluate a candidate summary, and SUPERSEIV~\cite{huang2004automatic} which is an unsupervised technique to evaluate video summarization algorithms that perform frame ranking. VERT~\cite{li2010vert} on the other hand was inspired by BLEU in machine translation and ROUGE in text summarization. Other techniques include pixel-level distance between keyframes~\cite{khosla2013large}, objects of interest as an indicator of similarity~\cite{lee2012discovering} and precision-recall scores over key-frames selected by human annotators~\cite{gong2014diverse}. It is not surprising thus that~\cite{truong2007video} observed that researchers should at least reach a consensus on what are the best procedures and metrics for evaluating video abstracts. They concluded, that a detailed research that focuses exclusively on the evaluation of existing techniques would also be a valuable addition to the field. This is one of the aims of this work. More recently, computing overlap between reference and generated summaries has become the standard framework for video summary evaluation. However, all these methods which require comparison with ground truth summaries suffer from the challenges highlghted earlier. Yeung et al. observed that visual (dis)similarity need not mean semantic (dis)similarity and hence proposed a text based approach of evaluation called VideoSet. The candidate summary is converted to text and its similarity is computed with a ground truth textual summary. That text is better equipped at capturing higher level semantics has been acknowledged in the literature~\cite{plummer2017enhancing} and form the motivation behind our proposed evaluation measures. However, our measures are different in the sense that a summary is not converted to text domain before evaluating. Rather, how important its selections are, or how diverse its selections are, is computed from the rich textual annotations in VISIOCITY. This is similar in spirit to~\cite{sharghi2018improving}, but there it has been done only for egocentric videos. As noted by~\cite{otani2019rethinking} "the limited number of evaluation videos and annotations further magnify this ambiguity problem". Our VISIOCITY framework precisely hits the nail by not only offering a larger dataset but also in proposing a richer evaluation framework better equipped at dealing with this ambiguity.

\section{\model Dataset}
\label{sec:dataset}

\subsection{Videos} \textbf{VISIOCITY is a diverse collection of 67 videos spanning across six different categories: TV shows (\emph{Friends}) , sports (Soccer), surveillance, education (Tech-Talks), birthday videos and wedding videos.} The videos have an average duration of about 50 mins. Summary statistics for VISIOCITY are presented in Table 2. Publicly available Soccer, Friends, Techtalk, Birthday and Wedding videos were downloaded from internet. TV shows contains videos from a popular TV series \emph{Friends}. They are typically more aesthetic in nature and professionally shot and edited. In sports category, VISIOCITY contains Soccer videos. These videos typically have well defined events of interest like goals or penalty kicks and are very similar to each other in terms of the visual features. Under surveillance category, VISIOCITY covers diverse settings like indoor, outdoor, classroom, office and lobby. The videos were recorded using our own surveillance cameras. These videos are in general very long and are mostly from static continuously recording cameras. Under educational category, VISIOCITY has tech talk videos with static views or inset views or dynamic views. In personal videos category, VISIOCITY has birthdays and wedding videos. These videos are typically long and unedited. The videos are available to see and download from the project website at https://visiocity.github.io/

\begin{table}[ht]
\small{
\centering
\begin{tabular}{|l|c|c|c|c|}
\hline
Domain & \# Videos & Duration &  Total Duration \\
\hline\hline
Sports (Soccer) & 12 & (37,122,\textbf{64}) & 12.8 hours  \\
TVShows (Friends) & 12 & (22,26,\textbf{24}) & 4.8 hours \\
Surveillance & 12 & (22,63,\textbf{53}) & 10.6 hours \\
Educational & 11 & (15,122,\textbf{67}) & 12.28 hours \\ 
Personal Videos (Birthday) & 10 & (20,46,\textbf{30}) & 5 hours\\
Personal Videos (Wedding) & 10 & (40,68,\textbf{55}) & 9.2 hours \\ \hline
All & 67 & (26,75,\textbf{49}) & 54.68 \\
\hline
\end{tabular}
\label{tab2:ourdatasetstats}
\vspace{-8pt}
\caption{Key Statistics of VISIOCITY. Third Column is in minutes (min,max,avg)}}
\vspace{-8pt}
\end{table}

\subsection{Annotations} The ground truth in \model is not \emph{direct} in form of the user summaries, but \emph{indirect} in form of concepts marked for each snippet. Being at a higher level, indirect ground truth can be seen as a 'generator' of ground truth summaries and thus allows for multiple solutions (reference summaries) of different lengths with different desired characteristics and is easy to scale. It also makes the annotation process more objective and easier than asking the users to directly produce reference ground truth summaries. 

Concepts are carefully selected list of verbs and nouns based on the type of the video and are given importance ratings based on the knowledge of the particular domain. \textbf{The concepts are organized in categories instead of a long flat list.} Example categories include 'actor', 'entity', 'action', 'scene', 'number-of-people', etc. Categories provide a natural structuring to make the annotation process easier and also support for at least one level hierarchy of concepts for concept-driven summarization. 

\begin{figure}
    \centering
    \includegraphics[width=0.8\textwidth]{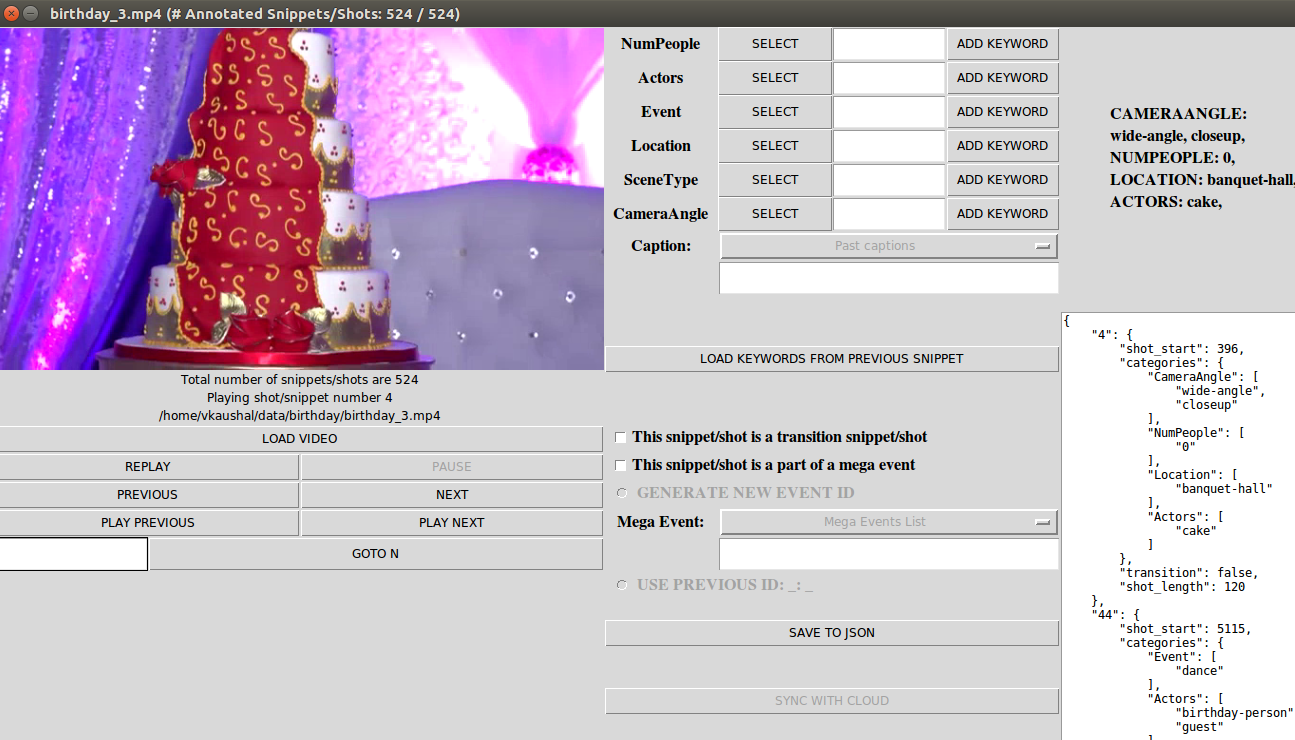}
    \caption{Annotation and visualization tool developed by us used in VISIOCITY framework}
    \label{fig:tool}
\end{figure}

In addition to concepts we ask annotators to group those consecutive snippets as \emph{mega-events} which together constitute a cohesive event. For example, a few snippets preceeding a goal in a soccer video, the goal snippet and a few snippets after the goal snippet together would constitute a 'mega-event'. A model trained to learn importance scores (only) would do well to pick up the 'goal' snippet. However, such a summary will not be very pleasing to watch because what is required in a summary in this case is not just the ball entering the goal post but the build up to this event and probably a few snippets as a followup. Thus this notion of mega events helps us to model the notion of continuity. 
\\
\\
\textbf{Textual Annotations vs Ratings or Scores as Indirect Ground truth:} While past work has made use of other forms of indirect ground truth like asking annotators to give a score or a rating to each shot~\cite{song2015tvsum}, using textual concept annotations offers several advantages. First, especially for long videos, it is easier and more accurate for annotators to mark all keywords applicable to a shot/snippet than for them to tax their brain and give a rating (especially when it is quite subjective and requires going back and forth over the video for considering what is {\it more important} or {\it less important}). Second, when annotators are asked to provide ratings, they often suffer from chronological bias. One work addresses this by showing the snippets to the annotators in random order~\cite{conf/cvpr/SongVSJ15} but it doesn't work for long videos because an annotator cannot remember all of these to be able to decide the relative importance of each. Third, the semantic content of a snippet is better captured through text~\cite{yeung2014videoset,plummer2017enhancing}. This is relevant from an 'importance' perspective as well as 'diversity' perspective. As noted earlier, two snippets may look visually different but could be semantically same and vice versa. Text captures the right level of semantics desired by video summarization. Also, when two snippets have same rating, it is not clear if they are semantically same or they are semantically different but equally important. Textual annotations brings out such similarities and dissimilarities more effectively. Fourth, as already noted, textual annotations make it easy to adapt VISIOCITY to a wide variety of problems.
\\
\\
\textbf{Annotation Protocol:} A group of 13 professional annotators were tasked to annotate videos (without listening to the audio) by marking all applicable keywords on a snippet/shot through a python GUI application developed by us for this task. It allows an annotator to go over the video unit by unit (shot/snippet) and select the applicable keywords using a simple and intuitive GUI (Figure~\ref{fig:tool}). It provides convenience features like copying the annotation from previous snippet, which comes in handy where there are are a lot of consecutive identical snippets, for example in surveillance videos. 

\textbf{Special caution was exercised to ensure high quality annotations.} Specifically, the guidelines and protocols were made as objective as possible, the annotators were trained through sample annotation tasks, and the annotation round was followed by two verification rounds where both precision (how accurate the annotations were) and recall (whether all events of interest and continuity information has been captured in the annotations) were verified by another set of annotators. Whatever inconsistencies or inaccuracies were found and could be automatically detected, were included in our automatic sanity checks which were run on all annotations. 

\section{Proposed Evaluation Framework}
\label{sec:eval}
Literature talks about certain desirable good characteristics of a video summary~\cite{gygli2015video, khosla2013large,lee2012discovering,ma2002user, zhang2016summary, zhou2018deep}. For example, a good video summary is supposed to be diverse (non-redundant), continuous or visually pleasing (without abrupt shot transitions), representative of the original video and contain important or interesting snippets from the video. In what follows, we dive deeper into these characteristics and propose measures to assess the candidate summaries on those characteristics.
\\
\\
\textbf{Diversity:} A summary which does good on diversity is non redundant. It contains segments quite \emph{different} from one another. \emph{different} could mean different things in terms of content alone (i.e. one doesn't want two similar looking snippets in a summary) or in terms of content and time (i.e. one doesn't want visually similar \emph{consecutive} snippets, but \emph{does} want visually similar snippets that are separated in time) or in terms of the concepts covered (one does not want too many snippets covering the same concept and would rather want a few of all concepts). In surveillance videos for example, one would like to have a summary which  doesn't  have  too many visually  similar  consecutive and hence redundant snippets,  but  does  have  visually  similar  snippets  that  are separated  in  time. For instance, consider a video showing a person entering her office at three different times of the day. Though all three look similar (and will have identical concept annotations as well), all are desired in the summary. With regards to the quantitative formulation, we define the first flavor of diversity as $$Div_{sim}(X) = \max \min_{i, j \in X} d_{ij}$$ where $X$ is a subset of snippets. $d_{ij}$ is IOU measure between snippets $i$ and $j$ based on their concept vectors. For the other two flavors of diversity, we define diversity clustered: $$Div(X) = \sum_{i=1}^{|C|} \max_{j \in X \cap C_i} r_j$$ where $C$ are the clusters, which can be defined over time ($Div_{time}$) (all consecutive similar snippets form a cluster) or concepts ($Div_{concept}$) (all snippets covering a concept belong to a cluster) and $r_j$ is the importance rating of a snippet $j$. When optimized, this function leads to the selection of the best snippet from each cluster. This can be easily extended to select a finite number of snippets from each cluster instead of the best one. 
\\
\\
\textbf{MegaEventContinuity:} element of continuity makes a summary pleasurable to watch. Since only a small number of snippets are to be included in a summary, some discontinuity in the summary is expected. However, the less the discontinuity at a semantic level, more pleasing is the summary to watch. There is a thin line between modelling redundancy and continuity when it comes to visual cues of frames. Some snippets might be redundant but are important to include in the summary from a continuity perspective. To model the continuity, VISIOCITY has the notion of mega-events as defined earlier. To ensure no redundancy \emph{within} a mega event, the mega-event annotations are as tight as possible, meaning they contain bare minimum snippets just enough to indicate the event. A non-mega event snippet is continuous enough to exist in the summary on its own and a mega event snippet needs other adjacent snippets to be included in the summary for semantic continuity. We measure mega-event continuity as follows 
$$MegaCont(X) = \sum_{i=1}^E r^{mega}(M_i) {|X \cap M_i|^2}$$ where, $E$ is the number of mega events in the video annotation, $r^{mega}(M_i)$ is the rating of the mega event $M_i$ and is equal to $\max_{\forall s \in M_i} r(s)$, $A$ is the annotation of video $V$, that is, a set of snippets such that each snippet $s$ has a set of keywords $K^{s}$ and information about mega event, $M$ is a set of all mega events such that each mega event $M_i$ ($i \in 1, 2, \cdots E$) is a set of snippets that constitute the mega event $M_i$
\\
\\
\textbf{Importance / Interestingness} - This is the most obvious characteristic of a good summary. For some domains like sports, there is a distinct importance of some snippets over other snippets (for eg. score changing events). This however is not applicable for some other domains like tech talks where there are few or no distinctly important events. With respect to the annotations available in VISIOCITY, importance of a shot or snippet is defined by the ratings of the keywords of a snippet. These ratings come from a mapping function which maps keywords to ratings for a domain. The ratings are defined from 0 to 10 with 10 rated keyword being the most important and 0 indicated an undesirable snippet. We assign ratings to keywords based on their importance to the domain and average frequency of occurence. Given the ratings of each keyword, rating of a snippet is defined as $r_s = 0$ if $\exists i: r_{K^s_i}=0$, and $r_s = \max_i r_{K^s_i}$ otherwise. Here $K^s$ is the set of keywords of a snippet $s$ and $r_{K^s_i}$ is the rating of a particular keyword $K^s_i$. Thus importance function can be defined as: $$\mbox{Imp}(X) = \sum_{s \in X \cap A \setminus M}r(s)$$ Note that when both importance and mega-event-continuity is measured, we define the importance only on the snippets which are non mega-events since the mega-event-continuity term above already takes care of the importance of the mega-event snippets.

As discussed earlier, since there are mutliple "right" answers with varying characteristics, we hypothesize that these are orthogonal characteristics and vary across different human summaries. For example, one human (good) summary could contain more important but less diverse segments while another human (good) summary could contain more diverse and less important segments depending on the intent behind summarization or user subjectivity. Also, in assessing summaries, one measure could be more relevant than another depending on the type of the video. For example in sports videos because of well defined events of interest, importance is more relevant in evaluating a summary. We verify our hypotheses experimentally. Motivated by this we propose using a suite of measures as defined above instead of overly depending on any one of them. The measures are computed using the annotations present in \model. We summarize them in Table~\ref{tab:measures}. We propose that a true and wholesome assessment of a candidate summary can only be done when this suite of measures (including the existing measures like F score) are used. Results and observations from our extensive experiments corroborate this fact.
\begin{table}[ht]
    \centering
    \begin{tabular}{|c|c|}
        \hline
        Measure & Expression \\
        \hline
        DiversitySim & $\max \min_{i, j \in X} d_{ij}$ \\
        Diversity(Time/Concept) &  $\sum_{i=1}^{|C|} \max_{j \in X \cap C_i} r_j$\\
        Mega Event Continuity & $\sum_{i=1}^E r^{mega}(M_i) {|X \cap M_i|^2}$\\
        Importance & $\sum_{s \in X \cap A \setminus M}r(s)$ \\
        \hline
    \end{tabular}
    \caption{Some of the proposed measures in \model. $X$ is the candidate summary, $C_i$s are clusters of consecutive similar snippets or concepts, $r$ denotes rating and $M$ denotes mega-events.}
    \vspace{-8pt}
    \label{tab:measures}
    \vspace{-8pt}
\end{table}

\section{Ground Truth Summaries for Supervised Learning}
\label{sec:gtsumm}
In practice it is difficult to acquire many human summaries with diverse characteristics especially for long videos. We explore strategies to automatically generate the reference ground truth summaries of desired lengths using the annotations present in \model. Specifcially, we use the above proposed assessment measures as scoring functions and maximize them to get the desired ground truth summaries. We note that maximizing a particular scoring function would yield a summary rich in that particular characteristic but it may fall-short on other characteristics. For example, a summary maximizing importance will capture the goals in a soccer video, but some snippets preceeding the goal and following the goal will not be in the summary and the summary will not be visually pleasing (example illustration at https://visiocity.github.io/). Hence a weighted mixture of such measures need to be maximized to arrive at optimal yet diverse reference summaries. This composite scoring function (weighted mixture) takes an annotated video ({\it keywords} and {\it mega-events} defined over snippets/shots) and generates a set of candidate ({\it ground-truth}) summaries which supervised or semi-supervised summarization algorithms can use. Mathematically, given $X$, a set of snippets of a video $V$, let $score(X)$ be defined as:

\begin{align*}
score(X, \Lambda) = \lambda_1 MegaCont(X) + \lambda_2 Imp(X) + \lambda_3 Div_{sim}(X) + \lambda_4 Div_{time}(X) + \lambda_5 Div_{concept}(X)
\end{align*}

This scoring function is parameterized on $\lambda$'s and is approximately optimized via a greedy algorithm~\cite{minoux1978accelerated} to arrive at the ground truth summaries. Different configuration of $\lambda$s generates different summaries. We explore two different strategies of identifying the right $\lambda$s for producing desired diverse reference summaries - i) pareto optimality, which is based on brute-force search  and (ii) proportional fairness, for which there is a known efficient greedy algorithm with a provable approximation guarantee for fairness.

\noindent \textbf{Pareto optimality}: 
Pareto optimality is a situation that cannot be modified so as to make any one individual or preference criterion better off without making at least one individual or preference criterion worse off. Beginning with a random element (a possible configuration of the lambdas) in the pareto-optimal set, we iterate over remaining elements to decide whether a new element should be added or old should be removed or new element should be discarded. This is decided on the basis of the performance on various measures. A configuration is better than another when it is better on all measures, otherwise it is not.

\noindent \textbf{Proportional Fairness}: Consider each configuration as an allocation to some agents (the different scoring terms) such that an allocation yields different performance on different measures (the value seen by the agents). Specifically, for a performance measure $p\in \{1,2,3,4,5\}$, an allocation/configuration $\Lambda_i$ yields  value $V_p(\Lambda_i)$. This setting allows us to use the notion of fair public decision making applying Nash social welfare equation~\cite{propfairness}, where the best allocation is one which is proportionally fair to all agents. We borrow from the approach in~\cite{propfairness} which studies a fairness property called \emph{core} that generalizes the notion of proportional fairness and pareto-optimality. Here, the problem is reduced to maximizing the  equation $F(\Lambda) = \sum_{p=1}^Nlog(1+V_p(\Lambda))$ for every configurations $\Lambda$ in the configuration space. Thus, computing the top $T$ configurations $\Lambda$ will take time $O(TNK)$. 

We verify experimentally that the automatic ground truth summaries so generated are at par with the human summaries both qualitatively and quantitatively. We use them in training the models tested on \model. 

\section{Towards A New State of the Art}
\label{sec:visiocity-sum}
Following~\cite{gygli2015video} we formulate the problem of automatic video summarization as a subset selection problem where a weighted mixture of set functions is maximized to produce an optimal summary for a desired budget. Specifically, given a video $V$ as a set of snippets $Y_v$ the problem reduces to picking $y$ $\subset$ $Y_v$ which maximizes our objective such that $|y| \leq k$, $k$ being the budget.\looseness-1
\begin{align} \label{max}
y^* = \operatorname*{argmax}_{y\subseteq Y_v, |y| \leq k} o(x_v, y)
\end{align}
$y^*$ is the predicted summary, $x_v$ the feature representation of the video snippets and $o(x_v, y)$ is the weighted mixture of components. \begin{align}
o(x_v, y) = w^Tf(x_v, y)
\end{align}
Our mixture model comprises of a submodular facility location term and modular importance terms. The facility location function is defined as  $f_{fl}(X) = \sum_{v \in V} \max_{x \in X} sim(v,x)$ where $v$ is an element from the ground set $V$ and $sim(v, x)$ measures the similarity between element v and element x. Facility-location thus models representativeness. The importance scores are taken from the VASNet model~\cite{fajtl2018summarizing} and the vsLSTM model~\cite{zhang2016summary} trained on VISIOICTY. The weights of the model are learnt using the large margin framework as described in~\cite{gygli2015video} using many automatic ground truth summaries and a margin loss which combines the feedback from different evaluation measures. Specifically, Given $N$ pairs of a video and an automatic reference summary $(V, y_{gt})$, we learn the weight vector $w$ by optimizing the following large-margin  formulation~\cite{taskar2005learning}:\looseness-1
\begin{align} \label{min}
\min\limits_{w \geq 0} \frac{1}{N} \sum_{n=1}^{N} L_n(w) + \frac{\lambda}{2}||w||^2 
\end{align}
where $L_n(w)$ is the generalized hinge loss of training example $n$ and $w$ is the weight vector.\looseness-1
\begin{align} \label{loss-aug}
L_n(w) = \max\limits_{y \subseteq Y_v^n} (w^T f(x_v^n, y) + l_n(y)) - w^T f(x_v^n, y_{gt}^n)
\end{align}
This objective is chosen so that each ground truth summary scores higher than any other summary by some margin. For training example $n$, the margin we chose is denoted by $l_n(y)$ and is a linear combination of the normalized losses reported by our proposed measures. We call our proposed method \model-SUM. We show that a simple model like this out-performs the current techniques (state of the art on TVSum and SumMe) on \model dataset because of learning from multiple ground truth summaries and learning from mutliple loss functions, each capturing different characteristics of a summary.

\section{Experiments and Results}
\label{sec:expresults}

\subsection{Implementation Details} 
For analysis of and comparison with human summaries, we generated 100 automatic summaries per video of about the same length as the human summaries. F1 score of any candidate summary is computed with respect to the human ground truth summaries following~\cite{zhang2016summary}. We report both avg and max. To calculate F1 scores of human summaries with respect to human summary, we compute max and avg in a leave-one-out fashion. 

For analysis of and comparison of different techniques on the VISIOCITY dataset, we report their F1 scores computed against the automatically generated summaries as a proxy for human summaries. We generate 100 automatic summaries for each video. All target summaries are generated such that their lengths are 1\% to 5\% of the video length. We test the performance of three different representative state of the art techniques on the VISIOCITY benchmark  vsLSTM \cite{zhang2016summary} is a supervised technique that uses BiLSTM to learn the variable length context in predicting important scores. It learns from a combined ground truth in terms of aggregated scores. VASNet~\cite{fajtl2018summarizing} is a supervised technique based on a simple attention based network without computationally intensive LSTMs and BiLSTMs. It learns from a combined ground truth in terms of aggregated scores and outputs a predicted score for each frame in the video. DR-DSN~\cite{zhou2018deep} is an unsupervised deep-reinforcement learning based model which learns from a combined diversity and representativeness reward on scores predicted by a BiLSTM decoder. It outputs predicted score for every frame of a video. To generate a candidate machine generated summary from the importance scores predicted by vsLSTM, VASNET and DR-DSN, we follow~\cite{zhang2016summary} to convert them into machine generated summary of desired length (max 5\% of original video). Our proposed model, VISIOCITY-SUM learns from multiple ground truth summaries and outputs a machine generated summary as a subset of snippets.

In all tables, AF1 refers to Avg F1 score, MF1 refers to Max F1 score (nearest neighbor score), IMP, MC, DT, DC and DSi refer to the importance score, mega-event continuity score, diversity-time score, diversity-concept score and diversity-similarity score respectively, as calculated by the proposed measures. All figures are in percentages. 

\subsection{Different human summaries have different characteristics} 
\begin{figure}
    \centering
    \includegraphics[width=\textwidth]{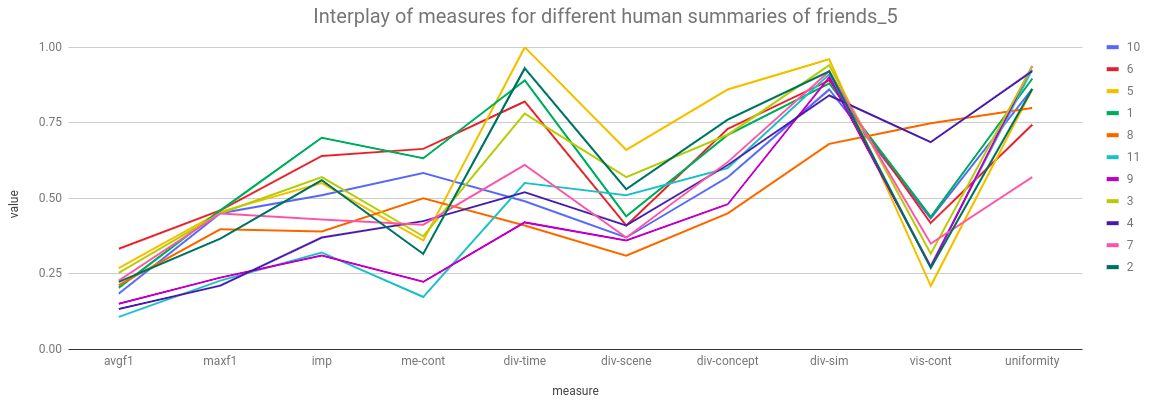}
    \caption{Different human summaries of same video perform differently on different measures}
    \label{fig:interplay}
\end{figure}

We asked a set of 11 users (different from the annotators) to create human summaries for two randomly sampled videos of each domain. The users were asked to look at the video without the audio and mark segments they feel should be included in the summary such that the length of the summary remains between 1\% to 5\% of the original video. The procedure followed was similar to that of SumMe~\cite{gygli2014creating}. We assess these human summaries qualitatively and quantitatively using the proposed set of performance measures and make the following observations. The human summaries are consistent with each other in as much as there are important scenes in the video, for example goals in Soccer videos. In the absence of such clear interesting events, the human summaries exhibit more inconsistency with each other. A representative plot (for the scores of human summaries of "friends\_5" video is presented in Figure~\ref{fig:interplay}). We note the following: a) proposed measures get good values on the human summaries as compared to uniform and random summaries, thus ascertaining their utility b) a human summary could score low on one measure and high on another measure c) the desired characteristics differ slightly across different domains (for example "importance" seems to be more important for soccer videos than diversity) 

\subsection{Automatically generated reference summaries are at par with human summaries}

In our experiments, we search for \emph{fair} configurations $\Lambda$ using both pareto-optimality and proportional fairness using the efficient greedy algorithm. Table~\ref{tab:auto-prop} shows the average F1 scores of both kinds of automatic ground truth summaries as compared to human summaries, uniform summaries and random summaries. We see that both approaches yield comparable performance. We use the automatic summaries generated using pareto optimality in the rest of our experiments.

\begin{table}[ht]
\begin{center}
\begin{tabular}{|l|r|r|r|r|r|r|} 
\hline
Domain &   Fri &  Soc &  Wed & Sur &  Tec &  Bir \\
\hline
Human & 24 & 30 & 21 & 35 & 20 & 21 \\
Uniform & 5 & 6 & 5 & 6 & 7 & 6 \\
Random & 6 & 5 & 5 & 6 & 6 & 6 \\
\emph{Auto-Pareto} & 25 & 27 & 14 & 31 & 25 & 17 \\
\emph{Auto-Prop} & 22 & 26 & 15 & 33 & 17 & 16\\ \hline
\end{tabular}
\end{center}
\caption{Performance (AF1) of human and auto summaries on videos across all the domains. \emph{auto-pareto} and \emph{auto-prop} stand for automatic summaries generated using pareto optimality and proportional fairness respectively.}
\label{tab:auto-prop}
\end{table}

\begin{figure*}
    \centering
    \includegraphics[width=\textwidth]{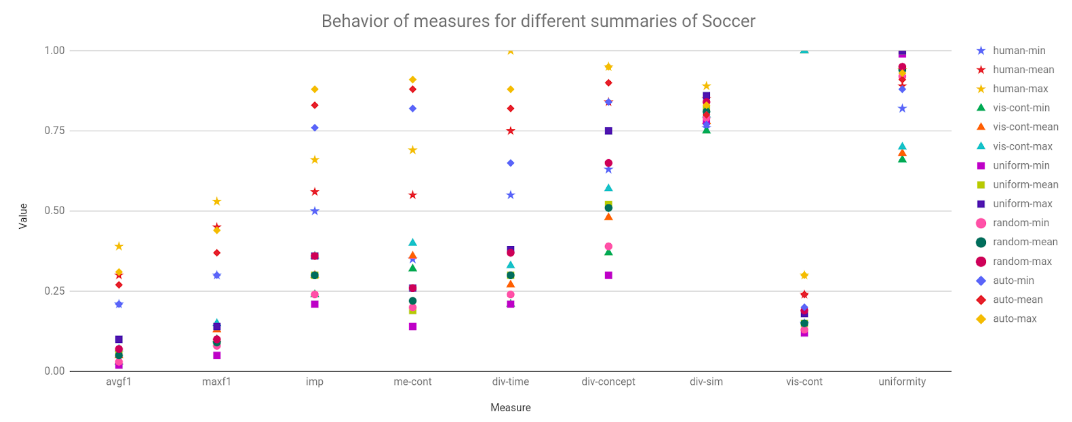}
    \caption{Behavior of different measures for different types of summaries for Soccer videos. 'vis-cont' summaries are visually continuous summaries assembled by picking a set of continuous snippets}
    \label{fig:measures}
\end{figure*}

We compare automatically generated reference summaries with human summaries on our proposed measures and present the quantitative results in Table~\ref{tab:humanauto}. We see that automatic and human summaries are much better than random on all the evaluation criteria. Next, we see that both the human and the automatic summaries are close to each other in terms of the F1 metric. The automatic summaries have the highest Importance, Continuity and Diversity scores. This is not surprising as they are obtained at the first place by optimizing a combination of these criteria. Figure~\ref{fig:measures} shows a representative plot for min, mean, max of different measures for different summaries of soccer videos. 

{
\small
\begin{table}[ht]
\centering
\begin{tabular}{|l|l|r|r|r|r|r|r|r|} 
\hline
Domain & Technique  & AF1 & MF1 & IMP & MC & DT & DC & DSi \\ \hline
\multirow{4}{4em}{Soccer} & Human & 30 & 45 & 56 & 55 & 75 & 84 & 85 \\
& Uniform & 6 & 9 & 30 & 19 & 30 & 52 & 82 \\
& Random & 5 & 9 & 30 & 22 & 30 & 51 & 81 \\
& Auto & 27 & 37 & 83 & 88 & 82 & 90 & 80 \\   \hline
\multirow{4}{4em}{Friends} & Human & 24 & 38 & 55 & 46 & 72 & 70 & 88 \\
& Uniform & 5 & 9 & 31 & 7 & 89 & 66 & 90 \\
& Random & 6 & 13 & 31 & 16 & 28 & 38 & 85 \\
& Auto & 25 & 41 & 87 & 69 & 76 & 85 & 81 \\
\hline
\multirow{4}{4em}{Surveillance} & Human & 35 & 56 & 58 & 65 & 45 & 79 & 80 \\
& Uniform & 6 & 8 & 12 & 9 & 12 & 49 & 55 \\
& Random & 6 & 8 & 13 & 12 & 13 & 46 & 55 \\
& Auto & 31 & 40 & 80 & 85 & 82 & 99 & 88 \\   \hline
\multirow{4}{4em}{TechTalk} & Human & 20 & 43 & 55 & - & 52 & 91 & 67 \\
& Uniform & 7 & 9 & 49 & - & 29 & 45 & 60 \\
& Random & 6 & 10 & 51 & - & 32 & 49 & 56 \\
& Auto & 25 & 43 & 86 & - & 78 & 93 & 96 \\   \hline
\multirow{4}{4em}{Birthday} & Human & 21 & 31 & 56 & 38 & 48 & 70 & 83 \\
& Uniform & 6 & 9 & 48 & 12 & 79 & 73 & 82 \\
& Random & 6 & 10 & 48 & 16 & 47 & 57 & 78 \\
& Auto & 17 & 30 & 86 & 81 & 63 & 91 & 84 \\   \hline
\multirow{4}{4em}{Wedding} & Human & 21 & 39 & 57 & 39 & 46 & 69 & 76 \\
& Uniform & 5 & 8 & 42 & 11 & 87 & 73 & 80 \\
& Random & 5 & 9 & 42 & 18 & 40 & 54 & 78 \\
& Auto & 14 & 21 & 81 & 79 & 71 & 95 & 88 \\   \hline
\end{tabular}
\caption{Performance of Human and Auto summaries for different domains. TechTalk videos do not have MegaEvents.}
\label{tab:humanauto}
\end{table}
}

\begin{figure*}
    \centering 
\begin{subfigure}{0.5\textwidth}
  \includegraphics[width=\linewidth]{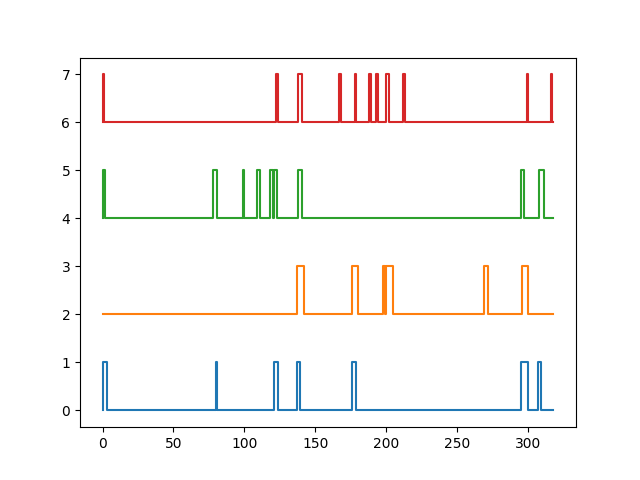}
  \end{subfigure}\hfil 
\begin{subfigure}{0.5\textwidth}
  \includegraphics[width=\linewidth]{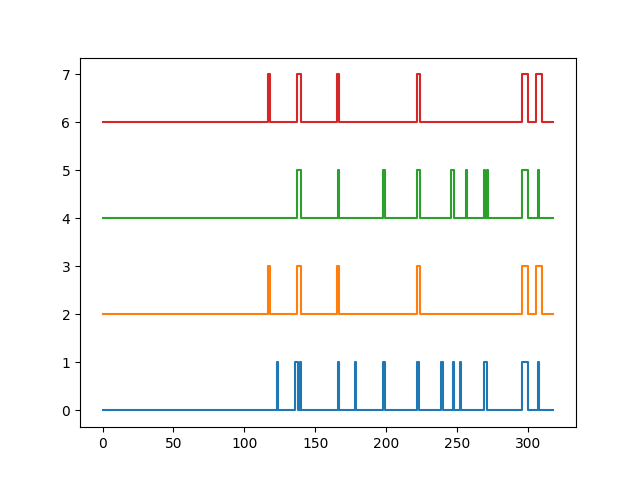}
\end{subfigure}
\caption{Shot numbers selected by some human summaries (left) and by some automatic ground truth summaries (right) for the friends\_5 video}
\label{fig:human-auto-qualitative}
\end{figure*}

We also compare the human and automatic summaries qualitatively. As an example, Figure~\ref{fig:human-auto-qualitative} compares the selections by human summaries (left) and automatic ground truth summaries (right) for \emph{friends\_5} video. We see a considerable similarity in selections, though a perfect match of selections is neither possible nor expected keeping with the spirit of multiple correct answers. Some human summary videos and automatic ground truth summary videos are reported at https://visiocity.github.io/. We see that a) it is very hard to distinguish the automatic summaries from human summaries and b) they form very good visual summaries in themselves. 

\subsection{VISIOCITY Benchmark: Performance of different models on \model}
We test some state of the art models and our simple enhancement \model-SUM on \model and report the numbers in Table~\ref{tab:results}.

{\small
\begin{table}[h!t]
\begin{center}
\begin{tabular}{|l|l|r|r|r|r|r|r|r|} 
\hline
Domain & Technique  & AF1 & MF1 & IMP & MC & DT & DC & DSi \\ \hline
\multirow{6}{4em}{Soccer} & Auto                          & 59.3                      & 93.3                      & 83.2                    & 84.3                        & 82.6                         & 85.9                            & 76.2                        \\
& DR-DSN                        & 2.8                       & 8.9                       & 23.7                    & 20.3                        & 23.2                         & 30.4                            & \textbf{83.4}                        \\
& VASNET                        & 28.4                      & 43.4                      & \textbf{63}             & 49.3                        & \textbf{62.1}                & \textbf{67.4}                   & 75.2                        \\
& vsLSTM                        & \textbf{31.9}             & \textbf{48.2}             & \textbf{62.2}           & \textbf{60.1}               & \textbf{62}                  & \textbf{69.5}                   & 76.5                        \\
& \textbf{Ours} & \textbf{32.6} & \textbf{50.3} & \textbf{64.2} & \textbf{62.6} & \textbf{63.4} & \textbf{72.2} & 78.7 \\
& Random & 3.4                       & 9.3                       & 25.7                    & 18.5                        & 25.5                         & 39.2                            & 80.5  \\ \hline
\multirow{6}{4em}{Friends} & AUTO & 66.3 & 96.9 & 87.8 & 84.6 & 80.3 & 89.8 & 83.1 \\
& DR-DSN & 4.3 & 9.4 & 19.1 & 6.9 & \textbf{65.7} & 51.5 & \textbf{98.5} \\
& VASNET & \textbf{17} & \textbf{29.6} & \textbf{41} & \textbf{39.3} & 49 & \textbf{60.6} & 86.7 \\
& vsLSTM & 15.5 & 27.2 & \textbf{40.4} & \textbf{39.2} & \textbf{64.7} & 59 & 91.1 \\
& \textbf{Ours} & \textbf{17.4} & \textbf{31.2} & \textbf{42.5} & \textbf{40.5} & 50.2 & \textbf{64} & 90.3 \\
& Random & 7.7 & 17.9 & 31.5 & 19.8 & 34.8 & 45.2 & 85.9 \\   \hline
\multirow{6}{4em}{Surveillance} & Auto & 62.4 & 96.8 & 81.8 & 83.2 & 78.6 & 98 & 85.2   \\
& DR-DSN & 10 & 17.7 & 33.6 & 20.2 & 21.8 & 54.5 & \textbf{57.2}   \\
& VASNET & \textbf{19.4} & \textbf{31.4} & \textbf{39.5} & \textbf{42.6} & \textbf{28.4} & \textbf{65.4} & 37.6   \\
& vsLSTM & 10.3 & 23.6 & 34.4 & 18.4 & 22.8 & 55.2 & \textbf{58.4}   \\
& \textbf{Ours} & \textbf{20.5} & \textbf{32.6} & \textbf{41.7} & \textbf{44.3} & \textbf{29.6} & \textbf{68.2} & 38.5 \\
& Random & 3.9 & 8 & 16.6 & 12 & 15.3 & 49.4 & 69.4   \\ \hline
\multirow{6}{4em}{TechTalk} & Auto & 64.7 & 91.5 & 79.8 & - & 80.5 & 88.4 & 94 \\
& DR-DSN & 13.5 & 22.5 & 49.3 & - & 24.8 & 29.9 & 35.2 \\
& VASNET & \textbf{18.2} & \textbf{35.7} & 52.1 & - & \textbf{47.3} & \textbf{43.3} & \textbf{43.2} \\
& vsLSTM & 15.1 & 32.2 & \textbf{60.3} & - & 38.8 & 35.3 & 41.7 \\
& \textbf{Ours} & \textbf{18.7} & \textbf{37.5} & 53.2 & - & \textbf{50} & \textbf{45.8} & \textbf{45.5} \\
& Random & 4.5 & 9.7 & 38.5 & - & 28 & 44 & 40.6 \\ \hline
\multirow{6}{4em}{Birthday} & Auto & 67.3 & 97.2 & 89.7 & 88.6 & 68.1 & 90.8 & 81.3 \\
& DR-DSN & 8.1 & 14.2 & 54.7 & 14.1 & \textbf{79.4} & 63.6 & \textbf{74.9} \\
& VASNET & 21.6 & 37.6 & 50.1 & 30 & 36.2 & 47 & 48.7 \\
& vsLSTM & \textbf{27.3} & \textbf{42.1} & \textbf{72.1} & \textbf{57.2} & 59.6 & \textbf{67.1} & 73.6 \\
& \textbf{Ours} & \textbf{28} & \textbf{44.3} & \textbf{74.8} & \textbf{60.3} & \textbf{62} & \textbf{69.5} & \textbf{77.6} \\
& Random & 6.9 & 14.2 & 51.8 & 16.9 & 49.2 & 54.8 & 70.3   \\ \hline
\multirow{6}{4em}{Wedding} & Auto & 55.4 & 94.4 & 83.9 & 74.7 & 67 & 88 & 85.7 \\
& DR-DSN & 4.2 & 8.9 & 40.7 & 14.4 & \textbf{76.6} & \textbf{62} & \textbf{88.4} \\
& VASNET & 4.5 & 14.4 & 46.5 & 22 & 44 & 52.7 & 84.9 \\
& vsLSTM & \textbf{9} & \textbf{17.3} & \textbf{50.2} & \textbf{29.5} & 50.1 & 56.9 & 80.7 \\
& \textbf{Ours} & \textbf{9.4} & \textbf{17.9} & \textbf{52.8} & \textbf{30.3} & \textbf{51.8} & \textbf{58.6} & 82.8 \\
& Random & 3.5 & 10 & 41.1 & 16.3 & 40.6 & 51.6 & 80   \\ \hline 
\end{tabular}
\end{center}
\caption{Comparison of different techniques on VISIOCITY. TechTalk videos do not have MegaEvents}
\label{tab:results}
\end{table}
}

We make the following observations: a) DR-DSN tries to generate a summary which is diverse. As we can see in the results, it almost always gets high score on the diversity term. Please note that the way we have defined these diversity measures, diversity-concept (DC) and diversity-time (DT) have an element of importance in them also. On the other hand, diversity-sim (DSi) is a pure diversity term where DR-DSN almost always excels. b) Due to this nature of DR-DSN, when it comes to videos where the interestingness stands out and importance clearly plays a more important role, DR-DSN doesnt perform well. In such scenarios, vsLSTM is seen to perform better, closely followed by VASNET. c) It is also interesting to note that while two techniques may yield similar scores on one measure, for example vsLSTM and VASNET for Soccer videos (Table~\ref{tab:results}), one of them, in this case vsLSTM, does better on mega-event continuity and produces a desirable characteristic in the summary. This further strengthens our claim of having a set of measures evaluating a technique or a summary rather than over dependence on one, which may not fully capture all desirable characteristics of good summaries. d) We also note that even though DR-DSN is an unsupervised technique, it is a state of the art technique when tested on tiny datasets like TVSum or SumMe, but when it comes to a large dataset like \model, with more challenging videos, it doesn't do well, especially on those domains where there are clearly identifiable important events for example in Soccer (goal, save, penalty etc.) and Birthday videos (cake cutting, etc.). In such cases, models like vsLSTM and VASNET perform better as they are geared towards learning importance. In contrast, since the interstingness level in videos like Surveillance and Friends is more spread out, DR-DSN does relatively well even without any supervision. e) \model-SUM does better than all techniques on account of learning from individual ground truth summaries and a combination of loss functions. 

\section{Conclusion}
In order to improve the objectivity and consistency in the design of video summarization benchmark datasets as well as their use in evaluating video summarization models, we present \model, a large benchmarking dataset and demonstrated its effectiveness in real world setting. To the best of our knowledge, it is the first of its kind in the scale, diversity and rich concept annotations. We introduce a recipe to automatically create ground truth summaries typically needed by the supervised techniques. Motivated by the fact that different good summaries have different characteristics and are not necessarily better or worse than the other, we propose an evaluation framework better geared at modeling human judgment through a suite of measures than having to overly depend on one measure. Finally we report the strengths and weaknesses of some representative state of the art techniques when tested on this new benchmark and demonstrate the effectiveness of our simple extension to a mixture model making use of individual ground truth summaries and a combination of loss functions. We hope our attempt to address the multiple issues currently surrounding video summarization as highlighted in this work, will help the community advance the state of the art in video summarization. We make \model available through the project page at https://visiocity.github.io/ and invite the researchers to test their algorithms on \model benchmark.

\section*{Acknowledgements}
This work is supported in part by the Ekal Fellowship (www.ekal.org) and National Center of Excellence in Technology for Internal Security, IIT Bombay (NCETIS, https://rnd.iitb.ac.in/node/101506)

\bibliographystyle{abbrv}
\bibliography{egb}

\end{document}